\pdfoutput=1

\documentclass[11pt]{article}

\usepackage[final]{EMNLP2023}

\usepackage{times}
\usepackage{latexsym}

\usepackage[T1]{fontenc}

\usepackage[utf8]{inputenc}

\usepackage{microtype}

\usepackage{inconsolata}

\usepackage{booktabs}
\usepackage{amsmath}
\usepackage{enumitem}
\usepackage{etoolbox}
\patchcmd{\quote}{\rightmargin}{\leftmargin 1em \rightmargin}{}{}

\title{``Is the Pope Catholic?'' Applying Chain-of-Thought Reasoning to Understanding Conversational Implicatures}

\author{Zae Myung Kim$^{1}$ \and David E. Taylor$^{2}$ \and Dongyeop Kang$^{1}$ \\
        Department of Computer Science and Engineering$^{1}$, Department of Philosophy$^{2}$ \\ University of Minnesota Twin Cities \\\texttt{\{kim01756,detaylor,dongyeop\}@umn.edu}
}

\begin{document}
\maketitle
\begin{abstract}
Conversational implicatures are pragmatic inferences that require listeners to deduce the intended meaning conveyed by a speaker from their explicit utterances. Although such inferential reasoning is fundamental to human communication, recent research indicates that large language models struggle to comprehend these implicatures as effectively as the average human. This paper demonstrates that by incorporating Grice's Four Maxims into the model through chain-of-thought prompting, we can significantly enhance its performance, surpassing even the average human performance on this task.
\end{abstract}

\section{Introduction}
Recent  large language models (LLMs) have shown impressive capabilities in tasks such as language generation, question answering, and machine translation \citep{bubeck2023sparks}.
One question that remains unanswered is whether these models understand the implications of conversation.

Conversational implicatures are a type of indirect meaning that is conveyed through language in a conversation, which is not explicitly stated.
For example, suppose John asks Mary, ``Are you coming to Paul's birthday party tonight?'' and Mary responds, ``I have a lot of work to catch up on.'' The point Mary is trying to convey to Paul is clear: Mary cannot come to Paul's party. What Mary literally says is that they [Mary] have a lot of work. But this is not (all) that Mary means. In particular, and given the specific context of this exchange (including common knowledge among its participants), what Mary \textit{conversationally implies} is that they will not be going to the party.

Conversational implications are present in every aspect of human communication. As LLMs are intended to faithfully model such communication, they must be able to understand and reason about conversational implications.

In this paper, we investigate whether LLMs like ChatGPT can comprehend conversational implicatures by conducting quantitative experiments (on \textsc{BIG-bench} and \textsc{LUDWIG} datasets) that ``prompt'' the model to give a binary (``yes'' or ``no'') answer to a given conversational scenario.
We find that while the model exhibits some basic understanding of these implicatures, its performance can be improved significantly when the reasoning process in computing the implicatures following Grice's Four Maxims \citep{grice1975logic} is demonstrated to the model in a step-by-step manner (i.e., ``chain-of-thought prompting'').

The paper is organized as follows:
In Section \ref{sec:understanding}, we first discuss what it means for an AI model to understand something.
Section \ref{sec:llms} briefly introduces what LLMs are and how they function to produce generations.
Section \ref{sec:grice} illustrates the quantitative experiments and the results.
Lastly, in Section \ref{sec:reflections}, we reflect on the findings and discuss their implications on the nature of conversational implicatures.

\section{Philosophical Perspectives on Understanding}\label{sec:understanding}
Our question ``do large language models understand conversational implicatures?'' presupposes an answer to a more fundamental question, namely, ``What is it for an AI to understand something in the first place?'' 
The purpose of this section is to provide an (operational) answer to the latter question in order to investigate the former. We begin with a very brief overview of the philosophical context that informs our conception of ``understanding.''

A central question in the philosophy of mind and cognitive science is whether genuine understanding--that is, understanding of the kind characteristic of human cognition--requires a form of subjective experience or consciousness that is distinct from the purely computational processes that underlie AI models.

To help frame the issue, imagine yourself in a room with a book of instructions for manipulating a set of symbols. Symbols come in; you match those symbols to the corresponding instructions; and you output a new string of symbols accordingly. Now suppose that the foreign symbols are literally symbols of a distant natural language L, one that you are not otherwise familiar with. If the instructions are good enough, and if you are good enough at following them, then the pattern of inputs and outputs that you manifest from inside the room will largely match that of a native L-speaker. Given this setup, the  question is: do you--in the room by yourself with the instructions--understand language L?

Philosophers like John Searle think that the answer is ``No.'' According to Searle, understanding is something over and above dispositions to linguistically behave in a certain way. What that extra something is a matter of empirical discovery, but it is arguably a ``biological phenomenon, and it is as likely to be as causally dependent on the specific biochemistry of its origins as lactation, photosynthesis, or any other biological phenomena.'' \citep{searle1980minds}

But not everyone is convinced by Searle's line of reasoning. Detractors point to facts that, first, AI models are designed to process and represent information in ways that are very similar to the way humans do and, second, such models can generate responses to linguistic stimuli that are contextually appropriate and coherent.

The classic version of this broad behaviorist approach to human cognition is the Turing test \citep{turing1950computing}, according to which a system exhibits genuine understanding if and only if, in theory, its responses are indistinguishable from those one would expect from a (statistically normal) human conversant.

It is the behaviorist approach, and the Turing test in particular, that we will adopt for the purposes of this paper. The primary rationale for this is methodological: in order to empirically measure the ability of an AI system to understand a natural linguistic phenomenon like implicature, we need a relatively precise empirical operationalization of understanding, which is precisely what the Turing Test gives us.

Therefore, in this paper, we will assume that a language model understands conversational implicatures if it generates a response that is deemed by human judges to match the appropriate response in a given conversational scenario.
This approach aligns with the behaviorist perspective, emphasizing observable behavior and measurable benchmarks.

\section{Large Language Models}\label{sec:llms}
LLMs are a type of artificial intelligence system that is designed to process natural language text, understand the context of the text, and generate appropriate responses.
These models are trained on vast amounts of text data, typically following next-token prediction task where the objective is to approximate the joint probability of some sequences of words occurring in a large corpus.
Formally, let $y_1,y_2,...,y_n$ denote tokens in a sentence, and $P(y_1, y_2,...,y_n)$ the probability of seeing all these tokens in this order.
Using the product rule of probability (i.e., the chain rule), we can decompose the probability of a text into conditional probabilities of each token given the previous ones:
\begin{align*}
&P(y_1, y_2,...,y_n) \\
&=  P(y_1) \cdot P(y_2|y_1) \cdot\cdot\cdot P(y_n|y_1,...,y_{n-1}) \\
&= \prod_{t=1}^n P(y_t|y_{<t}).
\end{align*}

So fundamentally, LLMs are nothing more than a probabilistic model that predicts the next most probable token that follows the given context.
Therefore, when we let the model generate after the sentence, ``What is the capital of South Korea?'' it is quite possible that the model will continue generating questions like, ``What is South Korea's largest city? What is the currency of South Korea?'' and so on. And this is because articles on the Internet (which the model happened to be trained on) could plausibly list out trivia questions about South Korea.

While a well-trained language model has the ability to recognize patterns and relationships between words and phrases that appear in diverse contexts, its text generation solely based on the most probable tokens may not align well with what humans consider appropriate.
To reduce this disparity, recent techniques, such as instruction tuning, have been proposed to fine-tune a pre-trained language model to perform specific tasks \citep{ouyang2022training}.
This technique involves providing the language model with a prompt (or an instruction), which is a specific text input that guides the model toward generating a desired output.
So when asked about the capital of South Korea, the instruction-tuned LLM will be likely to generate a response, ``The capital of South Korea is Seoul.''

Furthermore, instruction tuning with human preferences \citep{ziegler2020finetuning, ouyang2022training} has been highly effective in improving the accuracy and relevance of the outputs generated by language models, making them more useful for a wide range of applications.
In this paper, we aim to investigate the capacity of language models to comprehend conversational implicatures, with the goal of shedding light on the computational nature of these implicatures.

\section{Gricean Experiments}\label{sec:grice}
\subsection{Conversational Implicatures}
Conversational implicatures are a type of implied meaning that arises from a speaker's use of language in a particular conversational context. The standard theory of such implicatures comes from philosopher H. P. Grice. 
The theory is built on the idea that conversation is, at bottom, a goal-driven, rational activity. The goal may vary from conversation to conversation.
However, to the extent that one is involved in a given conversation, one will try to pursue that conversation's goal along with other participants \citep{grice1975logic}.

Specifically, Grice identified four maxims that guide this cooperative behavior of conversation: 
\begin{itemize}[noitemsep]
    \item \textbf{Quality}: only say what you reasonably believe to be true.
    \item \textbf{Quantity}: only provide information that is needed to achieve the purpose of the conversation.
    \item \textbf{Relevance}: make your contribution relevant to the conversation and its purpose.
    \item \textbf{Manner}: use relatively clear and concise language, within reason.
\end{itemize}

Conversational implicatures arise on certain occasions when a person appears to break one of these four maxims. Suppose for example that speakers A and B are both fully aware that the weather in B's location is particular unpleasant this time of year.

\begin{quote}
A: How's the weather over there?

B: Delightful, just as predicted!
\end{quote}

What B literally says is that the weather is delightful; but what they mean--what they conversationally imply--is that the weather is unpleasant, as expected.
The implicature arises because (i) speaker A has flouted the maxim of quality by saying something the falisity of which is common knowledge between A and B; (ii) the only way for B to reconcile what A says with A's presumed cooperation in the conversation is to suppose that what A really means is not what they say but rather something like ``the weather is terrible.''

Gricean conversational implicatures are important for understanding the nuanced and context-dependent nature of language use in communication. They also present a challenge for language models, which must be able to infer the intended meaning of a speaker based on the context and any implicatures that arise.

\subsection{Previous Studies and Datasets}
\citet{zheng-etal-2021-grice} were the first to look at the LLMs' capability to understand conversational implicatures by constructing a synthetic dataset called \textsc{GRICE}, highlighting the difficulty of the task for LLMs.
Also, similar efforts were made by researchers developing the ``Beyond the Imitation Game Benchmark (\textsc{BIG-bench}),'' \citep{srivastava2022imitation} which is a collaborative benchmark consisting of 200 tasks intended to probe large language models.
Comprehending conversational implicatures is one of those tasks.
More recently, \citet{ruis2022large} approached the problem with a new dataset (\textsc{LUDWIG}) consisting of naturally occurring implicatures and experimented with recent instruction-tuned LLMs such as ChatGPT.
In this paper, we conducted experiments with the test sets from \textsc{BIG-bench} and \textsc{LUDWIG} datasets, which consist of 492 and 600 test cases (i.e., conversations), respectively.

\subsection{Experimental Setting}
Following \citet{ruis2022large}, we conducted experiments in \textit{zero-shot} and \textit{few-shot} prompting.
Additionally, in this paper, \textit{chain-of-thought (CoT)} prompting was also utilized to verify the calculability argument for conversational implicatures.

\paragraph{Zero-Shot Prompting} refers to generating natural language text using LLMs without providing any explicit training examples related to the specific task of concern.
In zero-shot prompting, the LLM is tasked with generating responses based on its pre-existing knowledge and context learned during training on a diverse range of tasks.
An example of a zero-shot prompt looks as follows:

\begin{quote}
Esther and Juan are engaging in a conversation.
Esther asked ``Does it bother you that your wife goes away on long business trips?'' and Juan responded, ``Absence makes the heart grow fonder.''
What does Juan mean by his response? Answer by either ``yes'' or ``no''
\end{quote}

\paragraph{Few-Shot Prompting} provides the LLM with a few related examples (called shots) in the prompt to steer the model to produce generations in the way we want it to do.
An example prompt is:

\begin{quote}
The following examples are coherent sentences:

Esther asked ``Have you found him yet?'' and Juan responded ``They're still looking'', which means ``no.''

Esther asked ``Are you having fun?'' and Juan responded ``Is the pope Catholic?'' which means ``yes.''

Similarly, please, finish the following sentence by either ``yes'' or ``no'':
Esther asked ``Does it bother you that your wife goes away on long business trips?'' and Juan responded ``Absence makes the heart grow fonder.'' which means
\end{quote}

\paragraph{Chain-of-Thought Prompting} takes one step further and explicitly demonstrates the inferential reasoning steps involved in completing the task.
The main idea here is that by explaining to LLMs the detailed reasoning process in the examples, they will also follow and show the reasoning process when answering the prompt, and this often leads to more accurate results.
In our case, following the calculability argument put forward by Grice, we demonstrate a detailed step-by-step inferential reasoning process in examples included in the prompt:

\begin{quote}
Esther and Juan are engaging in a conversation.
Esther asked ``Juan, are you going to Paul's party?'' and Juan responded ``I have to work late.''
What does Juan mean by his response? Answer by either ``yes'' or ``no''

Esther wants to know an answer to her question: ``Juan, are you going to Paul's party?''
Juan responds: ``I have to work late.''
Juan's response, in the literal sense, does not count as a direct answer to the question.
On its face, Juan appears to be talking about something irrelevant to the question.
Yet Esther has no reason to believe that Juan is opting out of the operation of the cooperative principle which assumes that participants in a conversation cooperate with each other and attempt to be truthful, informative, relevant, and clear in order to facilitate successful communication.
Juan must therefore intend for Esther to infer an answer from ``what was said'' and background knowledge.
What would be the relevant background knowledge in this situation?
It is the fact that work-related duties typically take precedence over temporally co-located social gatherings.
Juan must therefore intend for Esther to infer that he will not attend the party due to him having to work late.
Thus, Juan means ``no'' from his response.

Answer: No.

\textsc{[Another example for positive case is shown here in a similar fashion]}

Esther and Juan are engaging in a conversation.
Esther asked ``Does it bother you that your husband goes away on long business trips?'' and Juan responded ``Absence makes the heart grow fonder.''
What does Juan mean by his response? Answer by either ``yes'' or ``no''
\end{quote}

In this manner, we generated three sets of test prompts for each test conversation in the \textsc{BIG-bench} and \textsc{LUDWIG} datasets accordingly.

\begin{table*}[ht]
\centering
\begin{minipage}{.48\textwidth}
\centering
\caption{ChatGPT's performance on the \textsc{BIG-bench} test set.}
\label{tab:big-bench} 
\begin{tabular}{lrrrr}
\toprule
 & prec. & rec. & F1 & acc. \\
\midrule
zero-shot & 0.78& 0.68& 0.64& 0.68\\
few-shot & 0.80& 0.77& 0.76& 0.77\\
chain-of-thought & \textbf{0.91} & \textbf{0.91} & \textbf{0.91} & \textbf{0.91}\\
\bottomrule
\end{tabular}
\end{minipage}
\hfill
\begin{minipage}{.48\textwidth}
\centering
\caption{ChatGPT's performance on the \textsc{LUDWIG} test set.}
\label{tab:ludwig}
\begin{tabular}{lrrrr}
\toprule
 & prec. & rec. & F1 & acc. \\
\midrule
zero-shot & 0.74& 0.65& 0.60& 0.63\\
few-shot & 0.75& 0.72& 0.70& 0.71\\
chain-of-thought & \textbf{0.87} & \textbf{0.87} & \textbf{0.87} & \textbf{0.87}\\
\bottomrule
\end{tabular}
\end{minipage}
\end{table*}

\subsection{Experimental Results}

Table \ref{tab:big-bench} and \ref{tab:ludwig} show the precision, recall, F1, and accuracy scores of predicting binary answers for the three prompting scenarios using the ChatGPT (\textsc{text-davinci-003}) model released by OpenAI.
We can see that the performance increases from zero-shot to few-shot and to CoT settings.
It is impressive to note that the CoT reasoning drastically improves the performance, outperforming the average human scores of 86.2\% accuracy reported in \citet{ruis2022large}.

Indeed, via the CoT reasoning, we verify that the model was able to identify the missing background knowledge correctly in many cases:\footnote{More examples (and code for experiments) can be found at \url{https://github.com/minnesotanlp/conversational-implicatures}.}

\begin{quote}
Esther wants to know an answer to her question: ``Can he walk now?''
Juan responds: ``He is as strong as an ox.''
...
What would be the relevant background knowledge in this situation?
\textit{It is the fact that ``being as strong as an ox'' connotes exceptional physical strength, which is indicative of sufficient mobility to walk.}
Juan must therefore intend for Esther to infer that he is indeed able to walk now.
...
Answer: Yes.
\end{quote}

\section{Reflections}\label{sec:reflections}
As mentioned in Section \ref{sec:llms}, research in LLMs has seen a huge breakthrough with the development of instruction tuning with human feedback.
While pre-trained LLMs were already exposed to a vast amount of texts and thus capable of understanding diverse contexts, the notion of correct or preferred generations was missing.
And the instruction tuning filled that gap by nudging the LLMs towards the preferred direction.

\citet{ruis2022large} found that even the instruction-tuned LLMs still struggle with understanding conversational implicatures, falling behind the performance of average humans.
This is due to the pragmatic nature of these implicatures where additional missing background knowledge or assumptions must be inferred and this may not be apparent to the LLMs.
However, in this paper, we found that by detailing the step-by-step process for inferential reasoning, the LLMs became much better at the task, confirming the calculability argument for these implicatures.

Does the fact that these reasoning steps had to be demonstrated undermine the capability of LLMs?
Not necessarily so.
While it is true that the models may require some further fine-tuning to accurately respond in the zero-shot setting, the generated results obtained from the CoT setting demonstrated the model's ability to deduce the missing logical connection which led to much-improved performance, beating the average human scores.
Therefore, from a behaviorist perspective, we can say that the recent LLMs coupled with CoT prompting are capable of understanding conversational implicatures.

\bibliography{anthology,custom}
\bibliographystyle{acl_natbib}

\appendix

\end{document}